# Data-Driven Regular Expressions Evolution for Medical Text Classification Using Genetic Programming

Jiandong Liu
School of Computer Science
University of Nottingham Ningbo China
Ningbo, China
jiandong.liu@nottingham.edu.cn

Ruibin Bai
School of Computer Science
University of Nottingham Ningbo China
Ningbo, China
ruibin.bai@nottingham.edu.cn

Zheng Lu
School of Computer Science
University of Nottingham Ningbo China
Ningbo, China
zheng.lu@nottingham.edu.cn

Peiming Ge
Techonology Department
Ping An Health Cloud Company Limited China
Shanghai, China
gepeiming649@jk.cn

Uwe Aickelin
School of Computing and Information Systems
University of Melbourne
Melbourne, Australia
uwe.aickelin@unimelb.edu.au

Daoyun Liu
Techonology Department
Ping An Health Cloud Company Limited China
Shanghai, China
liudaoyun035@jk.cn

*Abstract*—In medical fields, text classification is one of the most important tasks that can significantly reduce human workload through structured information digitization and intelligent decision support. Despite the popularity of learning-based text classification techniques, it is hard for human to understand or manually fine-tune the classification results for better precision and recall, due to the black box nature of learning. This study proposes a novel regular expression-based text classification method making use of genetic programming (GP) approaches to evolve regular expressions that can classify a given medical text inquiry with satisfactory precision and recall while allow human to read the classifier and fine-tune accordingly if necessary. Given a seed population of regular expressions (can be randomly initialized or manually constructed by experts), our method evolves a population of regular expressions according to chosen fitness function, using a novel regular expression syntax and a series of carefully chosen reproduction operators. Our method is evaluated with real-life medical text inquiries from an online healthcare provider and shows promising performance. More importantly, our method generates classifiers that can be fully understood, checked and updated by medical doctors, which are fundamentally crucial for medical related practices.

## I. INTRODUCTION

Given imminent proliferation of medical data on the Internet and the increasing needs for online medical service of current society, applying AI algorithm in medical service has become one of the most active research topics in the last few decades. Text classification, as a problem in general being studied for many years, has been introduced to medical domain to improve service performance [1]–[5]. Classifying relevant medical data into clinical informative categories such as symptoms or disease could potentially vastly reduce human labor cost in medical services. The applications of medical text classification not only can help doctors improve service quality by providing more structured information, but also build the foundation for more advance application such as automated/intelligent diagnosis. During the past several decades, many Machine Learning (ML) based solutions have been proposed for text classification problems. For example, Support Vector Machine (SVM) has been used to classify text documents by maximizing the margin from positive and negative samples for each category, predicting whether the text belongs to the category or not [6]. To use SVM, a feature vector is extracted from text by bag-of-words (BOW) model. Although widely used, BOW only has limited semantic representation power. To improve this, many efforts have been spent to expand BOW representation with semantic information [7]. As a result, word embeddings are introduced to automatically learn semantic relation among words [8]. By using word embeddings, neural network based methods such as Convolutional Neural Network (CNN) and Recurrent Neural Network (RNN) have bring significant advance in text classification [9]–[12]. For example, operating at either words or character level, CNN process text raw data in a similar way of dealing with image. portrays text as a source of raw data similar to images. With help of convolutions of features, language patterns in text can be learned without explicitly analyzing semantics. Despite great advancement in terms of precision and recall, text classification still struggles to incorporate domain knowledge into classifiers for even better performance and more importantly better understandability. This is especially true for medical text classification. To illustrate, while requiring no expert modeling, hence easy to be adopted in any specific domain, ML based methods are often criticized about their "black box" nature, impenetrable and inexplicable workings process [13]. Specifically, classifiers based on ML methods cannot be interpreted and modified by human experts with their domain knowledge when classification performance is not satisfying. In contrast to ML methods, rule-based approaches attempt to simulate human-like manner to classify text. In other words, rule-based approaches are able to construct rules according to domain experts through techniques such as decision trees. Such expert system is a typical application of simulating human-like decision manner and has been adopted in many text process tasks since 1990 [14], [15]. While easy to understand with reasonable performance, rule-based approaches are considered to be highly tied to specific domain and therefore adapting to new domains usually requires reconstruction.

In recently years, ML based approaches represent an interesting solution as no human effort is required once a suitable set of text is labelled with predefined categories as training set. However, ML approaches are well known to be less interpretable than rule-based methods. While this is usually not a big drawback in many professional domains, in medical domain, text classification applications require not only very high requirements, but also high reliability, verifiability, and ability to be manually updated if necessary. In such scenarios, medical experts, usually physicians, prefer to be able to verify the working process using their own domain knowledge and do not blindly trust a "black boxes" system that sometimes do not provide necessary classification performance and even worse almost impossible to improve manually. As a result, in many cases rule-based approaches are preferred by these medical experts compared to ML based approaches, for its interpretability even though it requires additional human efforts for initial preparations and later system updates. Note that ML based methods also usually require human efforts from domain experts when training set is prepared. In recent years, there have been many deep learning-based algorithms applied to medical text classifications [16], [17] and their performance has reached a very high level statistically. But still, the "black box" nature of deep learning methods are not preferable for direct use in practice, due to their poor interpretability and lack of performance guarantees [1].

Among many rule-based approaches, regular expression is one of the most popular techniques to provide an interpretable solution to text classification [2]. Regular expression-based approach is a classic text classification technique that uses predefined pattern matchings to provide binary text classifications. The technique has been used in many Natural Language Processing (NLP) studies in medical domains [4], [18]. According to our best knowledge, in most of such studies, regular expressions are constructed manually by experts with domain knowledge. Such human efforts are very labor intensive and error-prone. In addition, those regular expressions are normally constructed by a group of domain experts instead of a single one. Hence, the resultant regular expressions often lack required standards and the quality of each individual one highly depends on the particular expert's experience and skills. To overcome this drawback, there have been a few research focus on automatic generation of regular expressions from big data [1], [2], [19], [20]. [1] proposed an iterative constructive heuristic to generate regular expressions which can serve as drafts for manual improvement. However, like many greedy heuristics, the quality of this constructive heuristics is not guaranteed. None of these studies can produce classification performance that is comparable to those from manually generated regular expressions. This is caused by the fact that the searching space for automatically generated regular expressions are enormous due to the huge amount of feature words and their combinations using various regular expression operators. Searching for the optimal regular expressions in the entire solution space is extremely time consuming if even possible.

Furthermore, most previous automatic regular expression generation methods are designed to increase accuracy instead of facilitating the interpretability of their output regular expressions. While regular expressions are meant to be readable, those automatic generated ones are often long and making little sense to human experts and very hard for further manual improvement. We argue that, in real world scenario, clinical text especially those narrative ones produced by patients usually contain typos, abbreviations and non-standard terminologies. Automatically generated regular expressions based on such data can often result in overfitting if without final fine tuning by human experts.

To address those problems, we study an automatic regular expression generation method to classify clinical text to support clinical diagnosis process with informative medical guidance. Our method bases on a Genetic Programming (GP) framework that is specifically modified to satisfy needs for evolving regular expressions from real world applications. Our model has been tested with large size real world clinical data and experiments prove that our model produces very satisfactory results in medical text classification task. More importantly, the regular expressions generated by our method are more interpretable and structured friendly for human experts fine-tuning and checking.

The main contributions of this paper are summarized as follows:

1) A data driven GP with controlled tree depth was proposed specifically to learn pattern of medical narrative texts aiming at both better classification performance and interpretability of solutions. Our work can be considered as a novel attempt that uses optimization algorithm to automatically learn regular expression from medical domain data and provides much guidance value in real world application.

2) A hybrid model combining machine learning and regular

expression approaches are proposed for better performance from both worlds. This hybrid model uses regular expression classifier as complementary part for machine learning algorithm to improve the overall classification performance. Compared with other regular expression learning approach, our model gives better performance in both precision and recall.

## II. RELATED WORK

As an active research topic, text classification has been studied for many years toward automatic learning approaches. Tasks such as classifying semantic text [21], clinical records [5], and emotion text [22] are well solved by widely used machine learning algorithms with promising performance. With advance of neural network-based model, various types of approaches are designed and applied in text classification tasks. CNN has shown its excellent learning capability in modeling sentence with high-order features [5], [10], [23], classifying sentiment [24] and semantic [25] sentences and inferencing medical knowledge [16]. In addition, RNN provides excellent frameworks to solve problems with insufficient training data [9], [11].

Although machine learning approaches have shown promising performance in text classification tasks, practitioners often criticize their lack of interpretability, which provides human with no way of fine-tuning. As mentioned in previous section, regular expressions are often adopted in real world scenario to generate solutions in the situation where interpretability is essential. In general regular expressions are mainly adopted in NLP tasks such as text classification [2], [20], text extraction [26], etc. In order to reduce labor cost in construction regular expressions, many works have been conducted to exploit various ways of automatic generation of regular expression solutions [27], [28]. While producing promising results, these works usually focus on data with limited complexity and most of their regular expression solutions are syntactically simple. In our opinion, such regular expression solutions are not capable to solve NLP tasks with complex data, such as those presented in this work. In addition, those works have not been thoroughly applied to specific domain with large amounts of real-world data [29]. Thus, the robustness and suitability of these techniques for solving real world problem is still questionable. In order to test suitability, attempts of learning regular expression from large size of real domain data were conducted in many domains including detecting spam emails [30] and HTML detection [31]. In medical research field, regular expression based approaches are applied to text extraction from clinical records [32], symptom classification [2], and semantic recognition [33]. While those applications in real world problems demonstrate that learning regular expression from large size of data is feasible, solutions remain in simple architecture and have limited learning capability in dealing with long sequence of text. The drawback of those learning based approaches call for systems that are sophisticated enough to cope with long text. Along with the need for more sophisticated learning models, the solutions should be complex enough to find hidden pattern in training data. However, such models make output solutions much harder to be understood by human, which leave no easy way for further fine tuning.

Many efforts have been invested to learning regular expressions to produce a better solution. Those works mainly include constructive heuristic [1], [27], [34] and optimization methods [19], [20], [26] . To provide better performance, the process of our framework takes the advantage of GP [35], which employ the Darwinian principle of evolution to find a possible solution without the need of any prior information. Usually, the applications of GP in classification use only precision and recall as evolving guidance as they are the main objectives. Obviously, such traditional GP is limited by the fact that it takes no other consideration into account. We argue that in order to generate interpretable solutions, our technique has to integrate interpretability as evolving guidance into GP system.

The main contribution of our research is a novel framework of automatic learning regular expression via GP based method to compose effective and interpretable regular expressions. The approach not only achieves competitive performance with added interpretability, but also serves as a complementary model of machine learning approaches to correct their misclassified results.

## III. THE PROBLEM AND SOLUTION ENCODING

In this work, we proposed a novel framework to learn regular expressions for text classification from a large amount of data. The training data are manually labeled with a set of pre-defined categories by human experts. The following sections describe the notations and data samples used in this paper.

### A. Problem description

The problem to be solved in this paper is a text classification task. Given a set of labeled inquires $Q$ and a group of pre-defined categories $C$, our model should classify each $q \in Q$ to one category $c \in C$. For each category $c$, a regular expression vector $R_c$ is developed to match all the inquiries belonging to this category. Therefore, the solution to the problem is a list of $n$ regular expressions vector:

$$(R_1, R_2, R_3, R_4, R_5, R_6 ...... R_n)$$

where vector $R_i$ is designed for category $i$'s classification and contains a number of regular expressions:

$$R_i = <r_i^1, r_i^2, ..., r_i^m>$$

To check whether a particular inquiry $q$ belongs to a category $c$, the corresponding regular expressions in $R_c$ are executed one by one sequentially. If the inquiry $q$ is matched by any regular expression in $R_c$, the inquiry $q$ is said to be in category $c$. Otherwise, the inquiry $q$ does not belong to the given category. For each regular expression $r_i$, the following structure is adopted

$$ri = < P\ \#\_\#N >$$

where *P* and *N* are respectively the positive and negative parts and they are combined by *NOT* function to form a regular expression. A given regular expression containing *NOT* function matches strings which are matched by positive part *P* and not matched by negative part *N*.

*B. Data*

We collected online clinic consultation data from our collaborator for both training and testing. In this paper a total of 4,634,742 Chinese text records are collected within two weeks time [1]. All collected records are labeled by medical experts from our collaborator. To standardize our framework, we use International Classification of Disease Ninth Revision (ICD-9) principle as labelling standards. Table 1 shows a sample of (translated) records that we collected.

TABLE I
A SAMPLE OF MEDIAL INQUIRY RECORDS WITH LABELS

| Inquires | Disease category |
|---|---|
| My calf hurts as soon as it gets cold. What's going on? | Pain in limb |
| Doctor, could you please have a look at the seriousness of eczema? | Skin Inflammation |
| Biphenyl double fat drop pill is gone, please prescribe for me. | Hepatopathy |
| Cannot sleep at night, too much dreams. | Insomnia |
| Always sleep talking, Feel stressed | Insomnia |
| Will I get pneumonia if I have had fever for one month? | Pneumonia |
| I have a fever and coughs. | Pneumonia |
| My girlfriend gets a cold and sneezes a lot. | Upper respiratory infection |
| I started burping 2 days ago, it cannot stop. | Adult indigestion |
| I feel full by eating just a little and could not digest properly | Adult indigestion |
| My right knee feels painful when I go upstairs. It doesn't hurt, when I walk. | Knee pain |
| My period hasn't come this month. I did surgery one week ago. | Induced abortion |
| I see a little blood on my underwear. | Abnormal vaginal bleeding |
| I have sex last night and saw blood on my underwear this morning | Postcoital vaginal bleeding |

In our framework, there are 776 medical categories predefined by experts. We sorted categories according to their sizes of inquiries and found that the largest 30 categories make up to about 50% of the total inquires in our data. Thus, in our experiment, we randomly select records from the top 30 categories to test our proposed methods. In order to analyze the impact of size of the training data set on the performance of the algorithm, we formed 3 sub data sets with different size (2 million, 0.5 million and 0.2 million) and randomly select 80% and 20% of each sub data set to form training set and testing set.

*C. Proposed regular expression syntax*

In order to generate regular expression that can be fine-tuned, our model aims at achieving good classifying performance as well as readability for human experts to understand and modify solutions. In this paper, we define a solution as interpretable only if it can be easily read and fine-tuned by experts. Thus, we proposed a novel regular expression syntax that guarantees good interpretability. In the default regular expression grammar, a regular expression is a character class or literals combined by operators. In general regular expression, there are 3 types of operators usually used: alternation, look around, quantifiers.

In this paper, we restricted our regular expressions to 3 news operators only on the basis of alternation and quantifiers. Other more sophisticated operators are not used in the automated expression generation but can be used by human experts during fine-tuning stage. The regular expressions from our system will contain only feature words which are connected by proposed functions only (see Table 2). In addition, Table 3 is the notations of words and expressions used to form regular expressions. With help of our newly introduced syntax, our model reduces search space for optimization and at same time improves interpretability of solutions.

TABLE II
THE LIST OF PERMITTABLE FUNCTIONS IN THE PROPOSED REGULAR EXPRESSION SYNTAX

| Symbol | Function | Description |
|---|---|---|
| OR | $OR(w_1, w_2, ...w_n)$ | Function to match text contain $w_1$ or $w_2$ or ... or $w_n$. The element of *OR* function can be functions or feature words |
| AD | $AD(e_1, e_2, \{a, b\})$ | Function to match text contain $e_1, e_2$ and $e_1$ is on the left of $e_2$. Distance between $e_1, e_2$ should lays in range $\{a, b\}$. $e_1, e_2$ are corresponding expression to each other. |
| NOT | $NOT(P, N)$ | Matched texts which are matched by positive part *P* and not matched by negative part *N*. negative part *N* is formed by a group of negative expression $N_i$ connected by OR function. Negative expression $N_i$ can be feature words or functions. |

TABLE III
THE NOTATIONS OF WORDS AND EXPRESSIONS IN PROPOSED REGULAR EXPRESSION SYNTAX

| Symbol | Descriptions |
|---|---|
| w | A feature word in word dictionary[a] extracted from training set |
| e | An expression which is formed by a group of feature words connected by OR function |

[a]Word dictionary will be explained later.

For the purpose of better readability of regular expression, the following constraints are also applied to each of the regular expressions $R_i$. Later on, we will show, through experiments, that these constraints do not affect the quality the solutions but help reduce the search space of our GP method significantly.

*1)* First layer structure of regular expression $R_i$ is *NOT* function which connects *P* and *N*

*2)* Positive part *P* is formed by two expressions which connected by *AD* function.

3) Negative part *N* is formed by a group of negative expression $N_i$ connected by *OR* function. $N_i$ can be a feature word or a function.

4) Expressions of *AD* function should not contain any other nested functions except for *OR* function

5) Function *OR* in positive part should not contain any other nested functions except for itself.

Figure 1 is an example of regular expression formed by feature words and proposed syntax function.

## IV. METHODOLOGY

In this section, we propose a data-driven genetic programming method to evolve regular expressions. Figure 2 illustrates the overall stages of this method, including pre-processing, feature extraction and solution searching.

### A. Pre-processing

In pre-processing, we employ a popular Chinese word segmentation method, Jieba [36], to segment input text into words (note that our data is in Chinese as described in the previous section). For a given category *c*, all its inquiries *q* in $Q_c$ are considered as positive set and the rest of inquiries (denoted by $Q_c^-$) are considered as negative set. $Q_c^-$ is the complementary set of $Q_c$ given the universe set *Q*. After segmentation of positive set and negative set, positive word dictionary $w_p$ and negative word dictionary $w_n$ are obtained, together with the frequency of each word over the whole data set.

### B. Feature extraction

Over-fitting is one of the common problems in machine learning when solving classification tasks, where model give well performance in training/validation data but poor performance in testing data. To solve this problem, we introduce a measurement, average word frequency [1], to quantify uniqueness of feature words. Average word frequency $f_w^c$ is defined as the frequency of a given word $w \in W$ existing in all inquiries in $Q_c$ divided by the number of sentences in inquiries belonging to $Q_c$. The measurement is calculated as below:

$$f_w^c = \frac{\sum_{q \in Q_c} f_{w,q}^c}{\sum_{q \in Q_c} l_q} \quad (1)$$

where $l_q$ is the number of sentences in inquiry *q* and $f_{w,q}^c$ is the number of times word $w \in W$ existing in an inquiry $q \in Q_c$. The average word frequency indicates how popular a given word is in each category. A pre-defined threshold will be used to select words whose average word frequency is higher than threshold to be included as feature words.

In addition to average word frequency, A co-occurrence matrix is also calculated. We argue that a regular expression with good interpretability should contain hidden feature pattern of text for a given category and human should be able to understand the matching objective by simply reading it. Thus, being able to extract hidden feature patterns from training data is vital in feature extraction process. In order to achieve that, we applied co-occurrence matrix that describes how words occur together because co-occurrence data help our framework to captures the relationship between feature words.

Formally, co-occurrence matrix in our framework describes the frequency of a pair of feature words existing in one inquiry in a given order. For each category c, two matrices are constructed for both positive and negative feature words dictionaries $w_p$, $w_n$ extracted from $Q_c$ and $Q_c^-$, respectively. The co-occurrence formula is defined as follows:

$$M(i,j) = \sum_{q \in Q_c} \begin{cases} 1 & \text{if } pos_q(i) < pos_q(j) \\ 0 & \text{Otherwise} \end{cases} \quad (2)$$

where i and j represent feature words in inquires set and $pos_q(i)$ denotes the position of a given word *i* in a given inquiry *q*.

### C. Solution searching

We use genetic programming (GP) to evolve regular expressions. The overview of the algorithm is given in Figure 3.

*1) Initial population:* The initial population can be given by human experts as guidance or randomly selected from positive words dictionary. In this study, categories are pre-defined by medical experts to represent groups of similar symptoms, which means the category names usually contain domain knowledge. In order to utilize the domain knowledge, we firstly form a regular expression by category names as shown in Table 4 for examples. The distance element of AD function is randomly generated among range (1,10), and it will evolve during mutation operation. The negative part of every initial individual is $w_n^1$, the most frequent word frequent word in negative word dictionary. Then, we use the top (N-n) frequent words in positive words dictionary to form the rest of population in the same way, where N is the pre-defined population size and n is the number of feature words in category names.

TABLE IV
EXAMPLE OF INITIAL INDIVIDUAL GENERATED BY CATEGORIES NAME

| Name of categories | Individuals generated by categories name |
| --- | --- |
| Flu consultation | NOT(AD( flu,flu,{0,8}),OR($w_n^1$)) <br> NOT(AD(consultation,consultation,{0,6}),OR($w_n^1$)) |
| Bitten by mammals | NOT(AD( bitten,bitten,{0,9}),OR($w_n^1$)) <br> NOT(AD(mammals,mammals,{0,4}),OR($w_n^1$)) |

In our study, each individual starts with only one regular expression, but we will keep inserting regular expressions to the existing one to increase individual's complexity (and more importantly the fitness) until the end of evolution. In our experiment, for every 500 generations, a regular expression is formed by a feature word randomly selected from positive word dictionary and most frequent word of negative dictionary and the regular expression will be inserted to each individual in population.

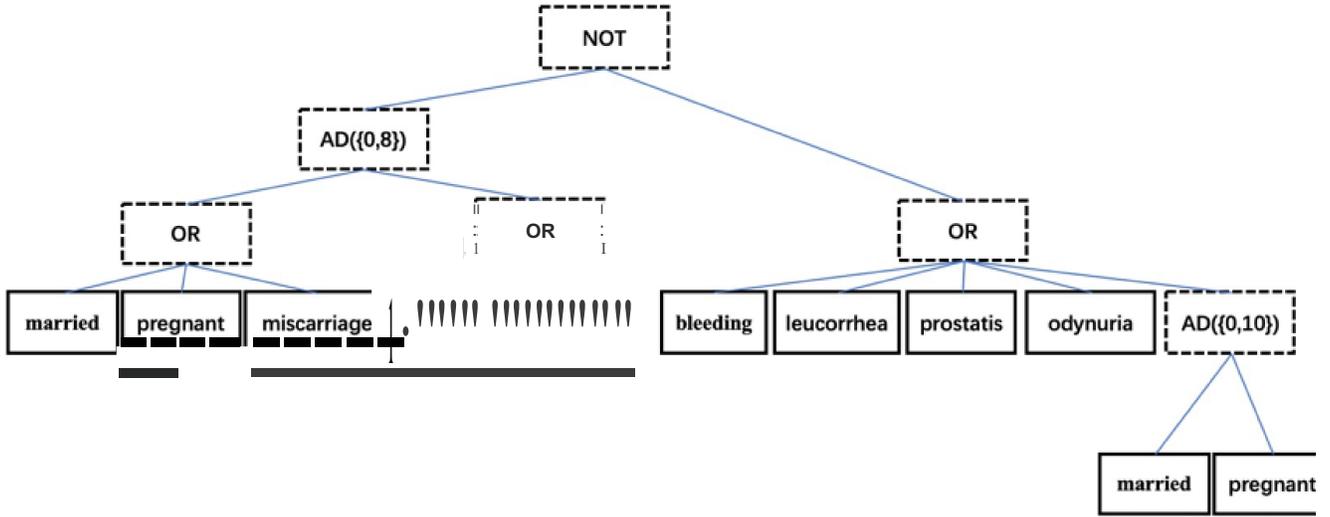

Fig. 1. An example GP tree of a regular expression

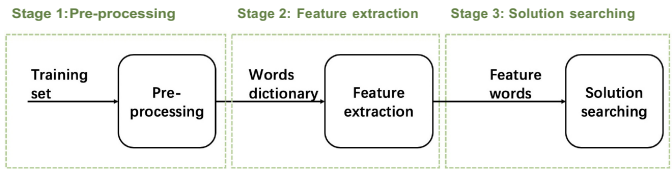

Fig. 2. Stages of proposed GP-based system

*2) Adaptive genetic operators:* After initial population is produced, genetic operators (crossover, mutation) are applied to all these individuals over generations. In this paper, we adopt single-point crossover and shrink mutation [37] methods to generate child individuals. we propose a self-adaptive genetic operation which can be described as follows:

$$P_c = S_{zg\text{-}rnoz}\left[\left(\frac{f_{avg}-f}{f_{ruu;-}f_{avg}}\right)\right] \qquad (3)$$

$$P_{rn} = S_{zg1noz}\left[a_{m}\left|f\frac{f_{avg}-f}{\cdot m\ ax}\frac{f}{avg}\right|\right] \qquad (4)$$

where $P_c$, $P_m$ are the probability of crossover, mutation operated on individual and $f$ is the fitness scores of individual. $f\ max$, $f_{avg}$ are the highest and average fitness scores in current population and $a_c$, $a_m$ are speed parameters of evolution. Those equations assign different probabilities of being operated to individuals according to their fitness via sigmoid function. Individuals with lower fitness will be assigned higher probability, which will accelerate evolution of low fitness individuals.

*3) Interpretability:* In our framework, a regular expression is considered interpretable when it is easy for doctors to read and fine-tune for better performance. Thus, an interpretable regular expression should be capable to learn hidden language pattern in a given category, and the feature words forming regular expression should be logically correlated. In order to find the hidden pattern, co-occurrence matrix(defined in Section IV B) is used. Co-occurrence matrix is used as guidance of selecting words in mutation operation occurred on positive part. According the proposed syntax and format constraints, positive part of a regular expression is formed by two group of expression combined by AD function. When mutating feature words in AD function, a feature word $w$ from expression $e$ will be replaced by another feature word $w$ selected from the word dictionary. Thus, calculating the correlation level of feature word $w$ towards the corresponding expression $e$ will be helpful in selecting words. In our work, we propose a measurement of correlation level as follow:

$$e = OR(w_1, w_2, \ldots, w_n)$$

$$C(w, e) = \sum_{i=1}^{n} \begin{cases} lvI(w, w_i) & \text{if } pos(e) < pos(e) \\ lvI(w_i, w) & \text{Otherwise} \end{cases} \qquad (5)$$

where $lvI(w, w_i)$ is the function that returns co-occurrence frequency of word $w$ towards words $w_i$. $pas(e)$ denotes the position of a given word expression $e$ in a given AD function. The correlation measurement function will be calculated for each word in positive word dictionary when the mutation is operated towards corresponding expressions. After obtaining every word's correlation towards expressions, we generate a probability of being selected via normalization as follows.

*positive word dictionary* $= (w_1, w_2 \ldots, w_m)$

$$probability(w) = \frac{C(w, e)}{\sum_{i=1}^{L7} C(w_i, e)} \qquad (6)$$

By doing so, our system integrates co-occurrence matrix into mutation operation to produce interpretable solution via expressing hidden language pattern.

*4) Fitness function:* Similar to other works, we use F score as only fitness function in our method. F score is the most popular performance metric for classification problems [38].

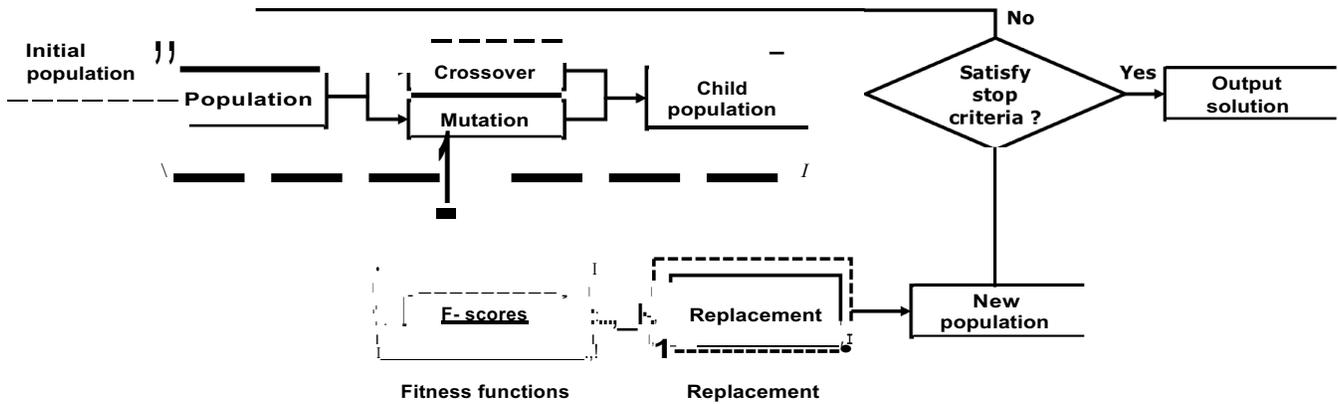

Fig. 3. Overview of the genetic programming algorithm for automated regular expression generation.

For a binary classification problem, F-Score is defined as follows:

$$Fe = \frac{(\beta^2 + 1) \cdot Precision \cdot Recall}{(\beta^2 \cdot Precision + Recall)} \quad (7)$$

where $\beta$ is the balancing parameter which decides preference between precision and recall.

  *5) Selection:* After offspring population is generated, parent and offspring population are combined to form new population and a replacement operator is applied on them to select $N$ individuals. In this paper, the individual with highest fitness score will be reserved and we randomly select $(N-1)$ individuals from the rest of population.

  *6) Stopping criterion:* The process of evolution continues until there is no change in the highest F-score of population for consecutive 100 generations. Then, the solution with highest F-score of last generation is returned as the final solution.

## V. EXPERIMENT

We conducted a comprehensive experiment to evaluate the performance of our algorithm on real-life data.

### A. Regular expression classifier

As discussed in previous part, the number of regular expressions that every solution contains depends on the stopping criteria and evolving process. In order to observe the impact of training size, we test the model on 3 different sizes of data set. Table 5 is the performances of our model on 3 data sets. As can be seen from the table, the larger size of training data leads to better performance of solution.

TABLE V
AVERAGE PERFORMANCE OF REGULAR EXPRESSION CLASSIFIER ON DIFFERENT DATA SET

| Training size | 200,000 | 500,000 | 2,000,000 |
| --- | --- | --- | --- |
| Precision | 0.8245 | 0.8649 | 0.9151 |
| Recall | 0.5642 | 0.6153 | 0.6353 |

Table 5 gives the average performance of the algorithm across 30 categories. The algorithm's performance varies among categories because of the diversity nature of different categories. For example, the algorithm obtains 96% precision and 91% recall in category *pregnancy preparation* whereas only achieve 85% precision and 43% recall in category *abnormal vaginal bleeding*. The standard deviation of precision and recall within categories are 6.3% and 18.7%, respectively. The reason behind such variation in terms of performance is largely due to the different concentration level of feature words in different categories. For category *pregnancy preparation*, feature words are clustered around relevant terms about *baby* and *mother*. In contrast, the feature words of category *abnormal vaginal bleeding* are relatively diversified, ranging from *sexual behavior, menstruation* to *endocrine disorder*.

### B. Solution Interpretability

Compared with machine learning based approaches, our framework provides an interpretable regular expression-based solution via interpretability guidance. The reason we integrate interpretability into evolving guidance is that our medical experts expect to fine-tune the regular expressions solutions when they cannot give satisfactory performance. Table 6 shows an example of regular expression solution fine-tuned by our medical experts.

### C. Performance enhancement by combining with deep learning methods

In addition to human fine tuning, we combine regular expression solutions with other machine learning model to improve classification performance. We selected Nai've Bayes, SVM, RNN and CNN as baselines and combined each of them with regular expression solutions. We applied the hybrid solution of machine learning and regular expression in our experiment. Specifically, if classification confidence of machine learning model is lower than 0.6, we employ regular expression classifier on the first 5 predictions in orders. The results are summarized in Table 7. Note that, due to limited computing resources, the results in Table 7 are based on 500,00 records data set.

TABLE VI
EXAMPLE OF FINE-TUNED REGULAR EXPRESSIONS

| Example of fine-tuned regular expressions |
|---|
| .*(pre-pregnancy\|pregnant\|expectant\|cyetic\|childing).{0,24}(examination\|test\| matters\|caution\|attention).* |
| .*(married\|pregnant).{0,8}(infertility\|examination). *# #.*(bleeding\| eucorrhea\|prostatis\|odynuria) .* |
| .*(follicle\|ovulation\| pregnancy).{0,30}(infertility).*'# #.*(vaginitis\| irregular menstruation) .* |
| Fine-tuned Regular Expression |
| .*(pre-pregnancy\|pregnant\|expectant \|cyetic\|childing\|family way\| *conception*\|ha.{0,8}bab).{0,24} (examination\|test\|matters\| caution\|attention). * |
| .*(married\|pregnant\|miscarriage). {0,8} (infertility\|examination). *#_#.* (bleeding\|leucorrhea\|prostatis\|odynuria\|period.{0,10}come).* |
| .*(follicle\|ovulation\| pregnancy).{0,30}(infertility).*# #.*(vaginitis\|irregular menstruation\|*irregular period*\|*menstruation.{0,10}irregular*\|period.{0,10} irregular).* |

TABLE VII
PERFORMANCE OF HYBRID CLASSIFIERS

| Solution | Precision | Recall |
|---|---|---|
| Naïve Bayes | 0.71 | 0.63 |
| Naïve Bayes + Regex | 0.82 | 0.68 |
| Naïve Bayes + fine-tuned Regex | 0.89 | 0.78 |
| SVM | 0.77 | 0.75 |
| SVM + Regex | 0.86 | 0.78 |
| SVM + fine-tuned Regex | 0.90 | 0.81 |
| RNN | 0.88 | 0.79 |
| RNN + Regex | 0.88 | 0.80 |
| RNN + fine-tuned Regex | 0.94 | 0.84 |
| CNN | 0.91 | 0.82 |
| CNN + Regex | 0.91 | 0.82 |
| CNN + fine-tuned Regex | 0.94 | 0.88 |

## VI. CONCLUSION

Although regular expression-based approaches have been used in text classification task for its robustness and easiness to understand, most of related works rely on manually constructing regular expressions which is a very labor consuming process. While reducing a lot of manual work, automatic regular expression generation often leads to suboptimal and less interpretable solution. In this paper, we propose a regular expression learning framework to solve text classification problem. In contrast to other regular expression learning approaches, we adopted a genetic programming optimization technique to generate interpretable solutions with help of novel regular expression syntax and GP encoding. In our case, solutions generated by the proposed framework not only solve the "black box" problem of current popular machine learning algorithms in text classification but also provide an easy way for doctors to modify regular expression classifiers when the performance is not satisfactory. We tested our algorithm on real-life medical text inquires and the experimental results show that the proposed algorithm is able to obtain good quality solutions. Although our framework cannot produce solutions which are comparable to machine learning solutions, it can be combined with machine learning approaches to form a hybrid model for even better performance.

In the future, we will systematically investigate in the impact of the proposed regular expression encoding on the solution quality and the search efficiency. We will also look into more efficient ways to combine GP based regular expressions with deep neural networks in addressing other interpretable text classification problems.